\documentclass[sn-mathphys]{sn-jnl}

\jyear{2021}%

\theoremstyle{thmstyleone}%
\theoremstyle{thmstyletwo}%

\theoremstyle{thmstylethree}%

\begin{document}

\title[Analysis and Adaptation of YOLOv4 for Object Detection in Aerial Images]{Analysis and Adaptation of YOLOv4 for Object Detection in Aerial Images}

\author[1,5]{\fnm{Aryaman Singh Samyal} }\email{aryamansinghsamyal@gmail.com}

\author*[2,5]{\fnm{Akshatha K R}}\email{akshatha.kr@manipal.edu}

\author[3,5]{\fnm{Soham Hans}}\email{soham.hans@gmail.com}
\author[4,5]{\fnm{Karunakar A K}}\email{karunakar.ak@manipal.edu}
\author[1,5]{\fnm{Satish Shenoy B}}\email{satish.shenoy@manipal.edu}
\affil*[1]{\orgdiv{Department of Aeronautical and Automobile Engineering}}

\affil[2]{\orgdiv{Department of Electronics and Communication Engineering, Center for Avionics}}

\affil[3]{\orgdiv{Department of Computer Science Engineering}}
\affil[4]{\orgdiv{Department of Computer Applications}}
\affil[5]{\orgdiv{Manipal Institute of Technology}, \orgname{Manipal Academy of Higher Education}, \orgaddress{\street{Manipal}, \postcode{576104}, \state{Karnataka}, \country{India}}}


\abstract{The recent and rapid growth in Unmanned Aerial Vehicles (UAVs) deployment for various computer vision tasks has paved the path for numerous opportunities to make them more effective and valuable. Object detection in aerial images is challenging due to variations in appearance, pose, and scale. Autonomous aerial flight systems with their inherited limited memory and computational power demand accurate and computationally efficient detection algorithms for real-time applications. Our work shows the adaptation of the popular YOLOv4 framework for predicting the objects and their locations in aerial images with high accuracy and inference speed. We utilized transfer learning for faster convergence of the model on the VisDrone DET aerial object detection dataset. The trained model resulted in a mean average precision (mAP) of 45.64\% with an inference speed reaching 8.7 FPS on the Tesla K80 GPU and was highly accurate in detecting truncated and occluded objects. We experimentally evaluated the impact of varying network resolution sizes and training epochs on the performance. A comparative study with several contemporary aerial object detectors proved that YOLOv4 performed better, implying a more suitable detection algorithm to incorporate on aerial platforms.}




\maketitle

\section{Introduction}\label{sec1}
There has been a rapid rise in aerial technology and systems over the past decade, especially in the domain of UAVs. They provide a large amount of visual data in videos, which opens possibilities for processing and extracting meaningful information from this data. Object detection capabilities can significantly increase the effectiveness of UAVs, which open up a multitude of applications such as automated flying and path planning, disaster management, weather forecasting, and military applications. Many challenges are faced during object detection and other forms of data extraction from aerial images and videos. Objects appear in various scales, making distant objects tiny and closer things much larger in perspective.
Objects either occluded or truncated are commonly seen in aerial images. Usually, there is a high density of object instances in each frame, making detecting each object even more difficult.  We used the VisDrone DET dataset \cite{zhu2020vision}  that mainly constitutes densely scattered, small objects. The viewpoint of the images in the dataset is highly variable. Sample UAV images highlighting these challenges are shown in Figure 1.
\begin{figure}[h!]
    \centering
    \includegraphics[width=0.85\linewidth]{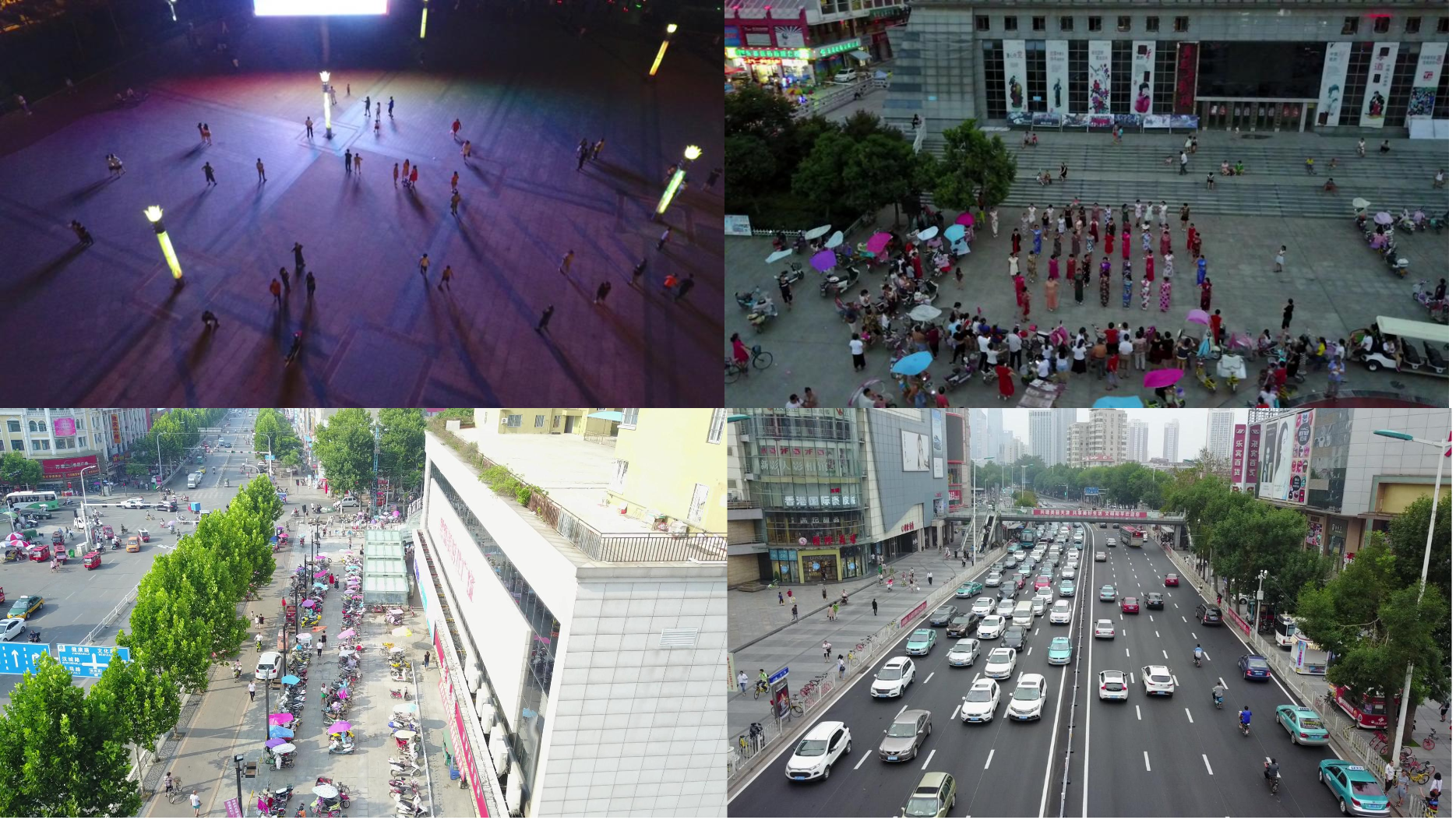} \vspace {-1mm}
    
    \caption{UAV captured aerial images from Visdrone dataset\cite{zhu2020vision}}\label{fig:1}
\end{figure}

 UAVs have limited memory and computational power; hence object detection algorithms need to balance between detection performance and speed to deploy for real-time requirements. Further, the computational inference cost should be minimal for increased feasibility and spatial flexibility, paramount when designing UAVs. The traditional approaches of using hand-crafted features with machine learning classifiers \cite{Dalal2005}, \cite{DPM2010}, \cite{Viola2004} are not capable enough to deal with the huge variations in the aerial images. Deep learning methods have proven to be extremely useful in generic object detection \cite{liu2020deep}, \cite{zhao2019object}. They mainly include the more accurate two-stage detectors \cite{lin2017feature}, \cite{he2015spatial}, \cite{fasterrcnn}, \cite{fastrcnn}, faster regression-based single-stage detectors \cite{yolo},\cite{redmon2017yolo9000}, \cite{yolov4}, \cite{liu2016ssd} and few anchorless based detectors \cite{FCOS}, \cite{centernet}.  However, all these methods focus primarily on object detection in datasets such as COCO \cite{coco} and PASCAL \cite{pascal}, where it aimed to detect objects of medium or large sizes. In this research work, we aim to apply a deep learning-based method for dealing with the challenges presented by aerial data. Many deep learning algorithms are not suitable for aerial object detection due to their incapability in detecting small-scale objects and large computation time.
\par YOLO \cite{yolo} and its various versions \cite{redmon2017yolo9000}, \cite{yolov4} have shown  excellent real-time detection performance in generic object detection due to its high computation speed. The YOLO algorithm works by dividing an input image into a grid of multiple cells. After that, the bounding boxes and their confidence scores are predicted for each cell, along with the class probabilities. The confidence is calculated as an intersection over union (IOU) metric, which measures the degree of overlap between the detected object and the ground truth as a fraction of the total area spanned by the two together (the union). In YOLOv2 \cite{redmon2017yolo9000}, anchor boxes were introduced, which are a pre-determined set of boxes. Therefore, the network predicts the offsets from these anchor boxes instead of directly predicting the bounding box. YOLOv3 \cite{redmon2018yolov3} included several small improvements, including the bounding box being predicted at different scales, Darknet being expanded to have 53 layers, and addition of objectness score to bounding box predictions. Alexey B. et al. introduced YOLOv4 \cite{yolov4} and performed a comparative study with current detectors; the result concluded that YOLOv4 was superior in terms of both speed and accuracy. YOLOv4 provided 43.5\% AP (65.7\% AP50 ) for the MS COCO dataset at a real-time speed of ~65 FPS on Tesla V100. YOLOv4 was twice as fast than EfficientDet with comparable performance. It improved YOLOv3's AP and FPS by 10\% and 12\%, respectively. Moreover, for drones, we require a balance between inference time and the accuracy of the detector. Further, the computational inference cost should be minimal for increased feasibility and spatial flexibility, which are paramount when designing UAVs. To fulfill these requirements, this research aims to analyze and show the adaptability of the single-stage YOLOv4 \cite{yolov4} detector when presented with the challenges of aerial images. The contribution of this research work is as follows. 
\begin{enumerate}
\item In this work, we have shown the adaptability of the popular YOLOv4 framework for aerial object detection tasks, using the transfer learning approach on VisDrone DET aerial object detection dataset.
\item The trained YOLOv4 model works well for high-density object instances and successfully detects truncated and occluded objects present in aerial frames.
\item  We have shown the significance of selecting varying network resolution sizes for detecting objects of ranging scales and perspectives.
\item  A performance comparison with the state-of-the-art aerial object detectors yielded that the proposed YOLOv4 object detector outperformed in terms of its detection speed, mAP, and other metrics.
\end{enumerate} 
\par The remainder of the paper is organized as follows. Section 2 briefs about the related work, and section 3 presents the method adopted in the paper. The experimental settings and results are reported in section 4, and section 5 discusses the results. Finally, section 6 concludes the paper with remarks.

\section{Literature Work}
Several deep learning approaches for generic object detection are modified to address aerial image object detection challenges and are listed in this section. Zhang et al. proposed a method for detecting small objects encountered in aerial view images \cite{zhang2019dense}. The paper introduced DeForm convolution layers and used the interleaved cascade architecture to detect dense and occluded objects. However, this model is a two-stage detector which leads to low operating speeds. 
Chen et al. proposed Re-Regression Net (RRnet), in which they mixed up the anchor-free detectors with a re-regression module to construct the detector which performs better for images with densely packed objects \cite{chen2019rrnet}. The anchor-free based detector firstly generates the coarse boxes and then a re-regression module is applied on the coarse predictions to produce accurate bounding boxes. This two-step process leads to an increase in inference time, hence making it infeasible for real-time detection. 
Lin et al. proposed HawkNet to up-scale feature aggregation method to fully utilize multi-scale complementary information which tackles the problem of unequal information transfer level of the backbone network that occurs in Feature Pyramid Network \cite{lin2020novel}. A novel up-sampling method is suggested for handling the existing method's ineffectiveness in reconstructing high-resolution feature maps. Liu et al. proposed HRDNet that attempted to take full advantage of high-resolution images without increasing computation costs \cite{liu2020hrdnet}. The paper introduced Multi-Depth Image Pyramid Network (MD-IPN) for maintaining multiple position information in multiple-depth backbones and Multi-Scale Feature Pyramid Network (MS-FPN), which can fuse multi-scale feature groups generated by MD-IPN to reduce information imbalance between these multi-scale multi-level features. The multiple scales helped HRDNet achieve better results than the state-of-the-art accuracy on small objects. 
Jin and Lin proposed an adaptive anchor for fast object detection \cite{jin2019adaptive}. This method uses the property that images taken at the same height have a clear scale range. The researchers use this to determine a scale of predefined anchors that can reduce the scale search range and risk of overfitting. SlimYOLOv3 is a method that uses fewer trainable parameters and floating-point operations (FLOPs) in comparison with original YOLOv3 for real-time object detection on UAVs \cite{slim}. They performed channel-level sparsity of convolution layers and pruned the channels with less information to obtain a slim object detector.
\par Though several methods have been proposed in the literature for aerial object detection tasks, many of them are modifications on the two-stage or anchorless detector. They aim to achieve a higher mAP; their detection speed is significantly less for real-time usage. Although Slim YOLOv3 is a fast detector with a real-time detection speed, the detection performance for aerial objects needs further improvement. YOLOv4 has shown a good balance between both accuracy and speed in various applications \cite{yang2020visionbased}, \cite{rs12111857}. Therefore in this research, we performed an experimental study to analyze the performance of YOLOv4 for object detection in UAV captured aerial images.
\section{Methods}\label{sec:approach}

In this research work, we investigate the domain adaptability of the YOLOv4 algorithm by training it with UAV-captured VisDrone dataset images using a transfer learning approach. The following section briefs about YOLOv4 architectures and their advantages over other object detectors.

\subsection{YOLOv4 Overview}
YOLOv4 \cite{yolo} is a single-stage detector that classifies and effectively localizes the objects in an image in one pass as illustrated in Figure 2. It was released in April 2020 and it included several data augmentation techniques, pre and post-processing methods as well as minor model modifications \cite{yolov4}. YOLOv4 can be briefly summarized as follows.

\begin{figure}[h!]
    \centering
    \includegraphics[width=0.85\linewidth]{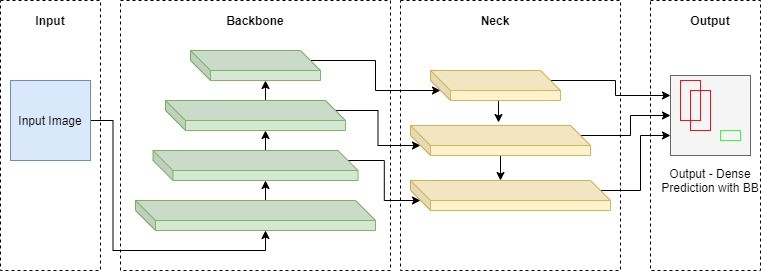} 
    \caption {YOLO: Single-stage architecture}

\end{figure}

\begin{itemize}
\item {Backbone - CSPDarknet53\cite{yolov4}:}: It is a modification of the Darknet-53 \cite{redmon2018yolov3} network utilized in YOLOv3. `53' denotes the number of convolutional layers present in the network. CSP implies the use of Cross-stage-partial connections as shown in Figure 3.
\item {Neck:} The neck consists of Path Aggregation Network (PAN) \cite{liu2018path} along with the implementation of Spatial Pyramid Pooling (SPP)\cite{he2015spatial}. The former performs as the method of parameter aggregation from different backbone levels for different detector levels, instead of the FPN (Feature Pyramid Network)\cite{lin2017feature} used in YOLOv3. SPP significantly increases the receptive field, separates the most significant context features, and causes almost no reduction of the network operation speed.
\item {Head:} YOLOv3 is used as the end of the chain object detector for dense predictions and detections.

\end{itemize}

\begin{figure}[h!]
    \centering
    \includegraphics[width=0.9\linewidth]{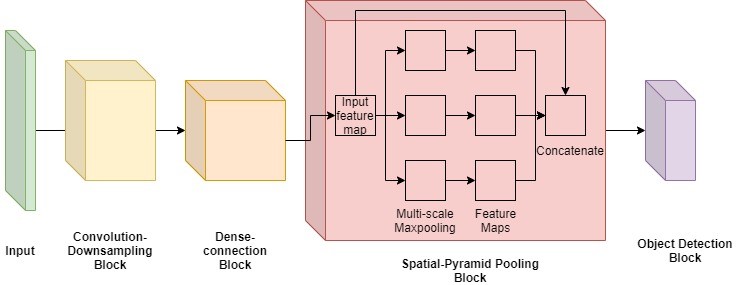} \vspace {-1mm}
    \caption{YOLO with SPP}
   
\end{figure}

In SPP shown in Figure 3, the input feature map is divided into d X d bins, following which maximum pool is implemented on each bin for each channel. For YOLO, SPP has been altered to preserve the output spatial dimension.

\par YOLOv4 is exceptionally efficient and suitable for aerial object detection due to the several innovative methods termed `bag of freebies' and `bag of specials' incorporated into the model. The use of CIoU loss function improves the model's convergence speed and accuracy by minimizing the central point difference between the predicted and ground truth bounding box. This allows YOLOv4 to provide accurate predictions, even on small objects. Moreover, the use of data augmentation techniques such as CutMix, Mosaic, and Self-adversarial training increases the robustness of the algorithm compared to previous YOLO versions. The implementation of methods like CSP, Multi-input weighted residual connections (MiWRC) and PAN, drastically reduces inference speed and computational cost, which are crucial for real-time object detection.
These features enable the detector to overcome challenges such as highly dense images and small object instances which are common in aerial datasets.
\subsection{Transfer Learning}
Transfer learning is a modern technique deployed for faster convergence while training deep learning algorithms. It comprises of utilizing a pre-trained model, trained on a different dataset (usually a dataset for general object detection, like MS COCO), for training on the actual task dataset that is relevant to our objective. Using this method, unless the dataset for the pre-trained model is significantly contrasting with final dataset, usually leads to the model being more robust and powerful. For our case, the model was loaded with the pre-trained weights only for layers responsible for low-level feature detection (first 137 layers of the model), thereby ensuring that layers responsible for feature concatenation and classification of higher-level features, is completely customized for our task, i.e., Aerial Object Detection. This was important as the MS COCO dataset does not have a lot of classes involving smaller object instance area, which is a common occurrence in aerial datasets. 

\subsection{Model Selection }
We have done extensive and detailed experimentation with YOLOv4 for aerial object detection. Our work included multiple combinations and modifications of the YOLOv4 network resolution to finalize the best result for our purpose. Furthermore, substantial testing was conducted to compare the performance of the model at various training iterations. During testing, multiple network resolution configurations of the model and AP and F1 score at varying iterations were analyzed to select the combination with the best performance. Our approach is mainly divided into two parts.
\begin{enumerate}
\item Checking various network sizes to select one with the best metric scores:
\par The YOLOv4 model was created using the darknet framework and was reconfigured to have a network resolution size of 768 * 768. This was done to get better performance (average precision) for detecting smaller objects, as the dataset had an abundance of such instances. The model configuration consisted of three YOLO layers. Further, mosaic data augmentation was enabled to improve ruggedness.  Although the model was trained with a network size of 768 * 768, we tested various network resolution configurations to select one with the highest performance. For the various models with different network sizes that were tested, no additional training was performed due to the limitation of resources, and hence, weights of the original model were used.
\item Finding at what number of iterations the model performed best:
\par The model's performance at various training iterations was analyzed to choose the best performing model. AP and F1 scores are the metrics used to compare the performance at various iterations.
\end{enumerate}
\section{Experiments and Results}
The entire experiment including training the model and evaluation was performed using Google Colab. The assigned GPU's by the cloud service included Tesla P100-16GB and Tesla K80, for different runtimes.
\subsection{Dataset Details}

The VisDrone2020-DET dataset \cite{zhu2020vision}, of the VisDrone Object Detection Challenge 2020 \cite{pailla2019visdrone} was utilized for training and testing. The dataset comprises of aerial images shot taken from a multirotor drone. This has varying scale objects due to different altitudes and camera angles of the drone while the images were captured. The perspective angle, occlusion, and illumination are notably diverse for different scenarios. Moreover, due to the common occurrence of truncated or occluded object instances in aerial images/frames. The `trainset' and the `valset' were combined to make the training dataset. This was done to increase the size and variability of the training data for a better and more robust model. The `trainset' and the `valset' consisted of 6,471 and 548 images, respectively, combining for a total of 7,019 images for the training dataset. The `testset-dev' was used as the validation/test set. The test set consisted of 1,610 images. The dataset mainly categorized into two parts - humans and transportation. There are two classes under humans, `Pedestrian' and `Person'. `Pedestrian' are humans standing or walking, whereas `Person' is humans with other poses. Transportation consists of eight classes of different vehicles.
 Thus, the dataset aggregates ten object classes as shown in Figure 4.
\begin{figure}[h!]
   \vspace {-4mm}
    \centering
    \includegraphics[width=0.85\linewidth]{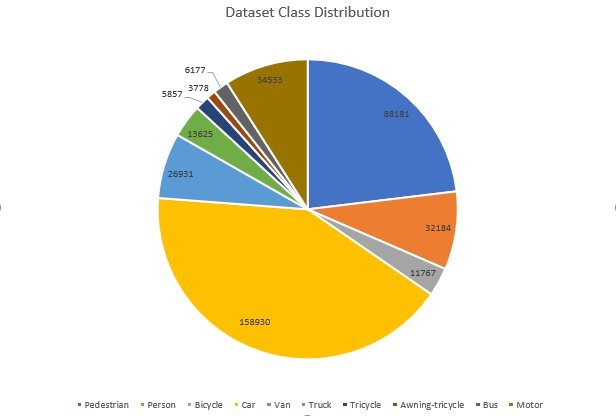} \vspace {-1mm}
    \caption{Dataset class distribution in VisDrone DET dataset}
    \begin{description}
                \item[Figure 4 Caption:] Dataset class distribution in VisDrone DET dataset
                \item[Figure 4 Alt Text:]A pie chart representing the number of instances for each class in the dataset. In descending order, it is as follows; Car, Pedestrian, Motor, Person, Van, Truck, Bicycle, Bus, Tricycle, Awning-Tricycle.    
         \end{description}
    
\end{figure}

The ground truth labels of the Visdrone dataset include annotation for truncation and occlusion of the objects present in an image. It has been incorporated for developing a model which does not detect objects void of the full feature set due to partial truncation. 
For reducing complexity and for developing a model which is pragmatic and more suitable for actual scenarios in aerial detection, we have not considered truncation and occlusion ratios as our feature inputs, and our model has been trained to detect such instances as well.

\subsection{Training Setup}
\begin{enumerate}
\item The training was initiated by loading pre-trained weights available on Alexey Bochkovskiy's GitHub repository \cite{alex} for YOLOv4 trained on MS COCO dataset \cite{coco} onto the first 137 layers of the framework, which are responsible for low-level feature detection. 

\item The random flag was set to 1 to increase the precision by training the model with different network sizes for every ten iterations.
\item Learning rate was set to 0.001, and `burn in' was set to 1000, which defines the number of batches after which the learning rate increases from 0 to the learning rate in epoch 0. Momentum and weight decay are set to 0.949 and 0.0005, respectively. These are the default suggested values for YOLOv4 to achieve optimal loss decay and convergence. 

\item Batch size and mini-batch size was set to 64 due to limitations set by the available GPU memory. Hence, for our training set, which consisted of 7,019 images, one epoch is given by 7019$/$64, which when rounded off to the nearest whole number gives 110 iterations.
\item The model is trained for 10,000 iterations, at which the loss and mAP had stagnated.
\end{enumerate}
\subsection{Experimental Results}
The experimentation performed included a comprehensive comparative study between various single-stage object detection and current baselines for selection of the optimal model for aerial detection. The results conclude that the final model was better than contemporary models and was ideal for our purpose.
\subsubsection{Model Selection:}
The experimental results obtained during model selection, as explained in methods, are shown below.
\subsubsection*{AP and F1 score versus iterations:} Figure 5 and Figure 6 shows the performance in terms of AP and F-1 score for various training iterations. It is found that the model achieved the highest AP and F1 scores for 8000 iterations. After that, the model utilized the weights saved at 8000 iterations for all metric calculations as given in the upcoming sections.

\begin{figure}[!h]
  \centering
  \begin{minipage}[c]{0.49\textwidth}
    \centering
    \includegraphics[width=0.99\linewidth]{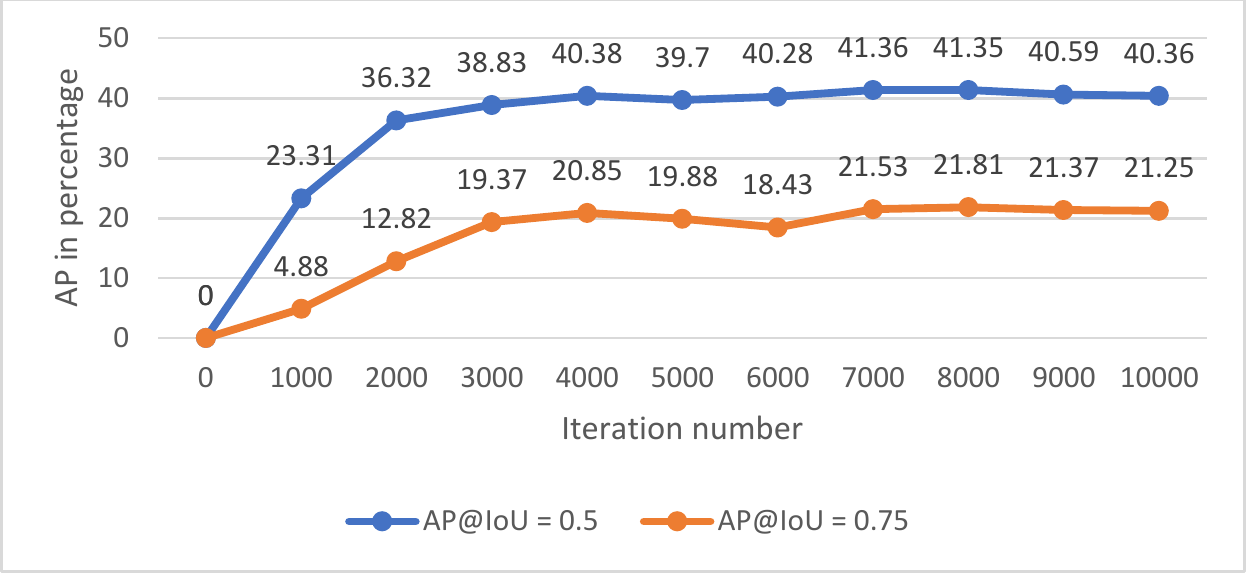}
    \caption{AP versus Iterations}

  \end{minipage}
  \hfill
  \begin{minipage}[c]{0.49\textwidth}
    \centering
    \includegraphics[width=0.99\linewidth]{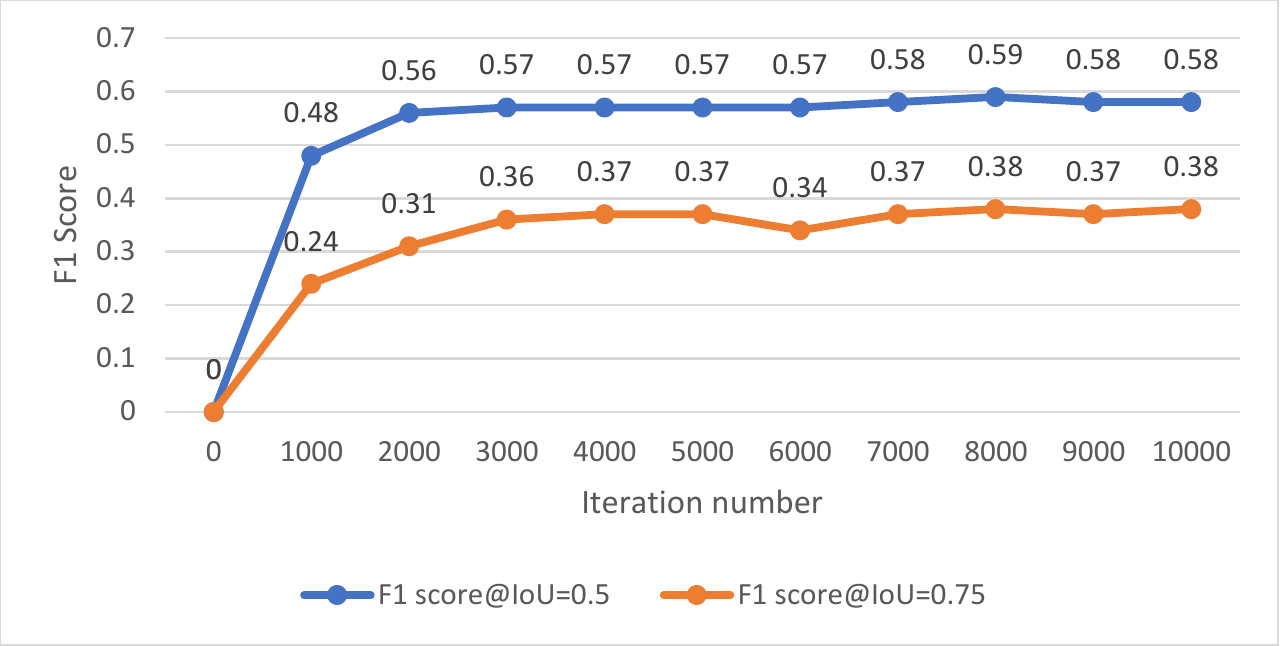}
    \caption{F-1 score versus iterations}

  \end{minipage}
\end{figure}

\subsubsection*{Network Resolution:}
 The model trained with 768 * 768 sized input images was tested with various network sizes to observe the effect of input resolution on detection performance. Figure 7 and Figure 8 shows the AP and detection speed for various network resolutions. Our results inferred that the model performed best in terms of AP with a network size of 1120 * 1120. The model with 416 * 416 resolution utilized a minimum inference time of 114 ms as observed in Figure 8. It was then concluded that for all metric calculations, weights of the trained model at 8000 iterations would be loaded on the model with a network size of 1120 * 1120.

\begin{figure}[h!]
  \centering
  \begin{minipage}[b]{0.49\textwidth}
    \centering
   \includegraphics[width=0.99\linewidth]{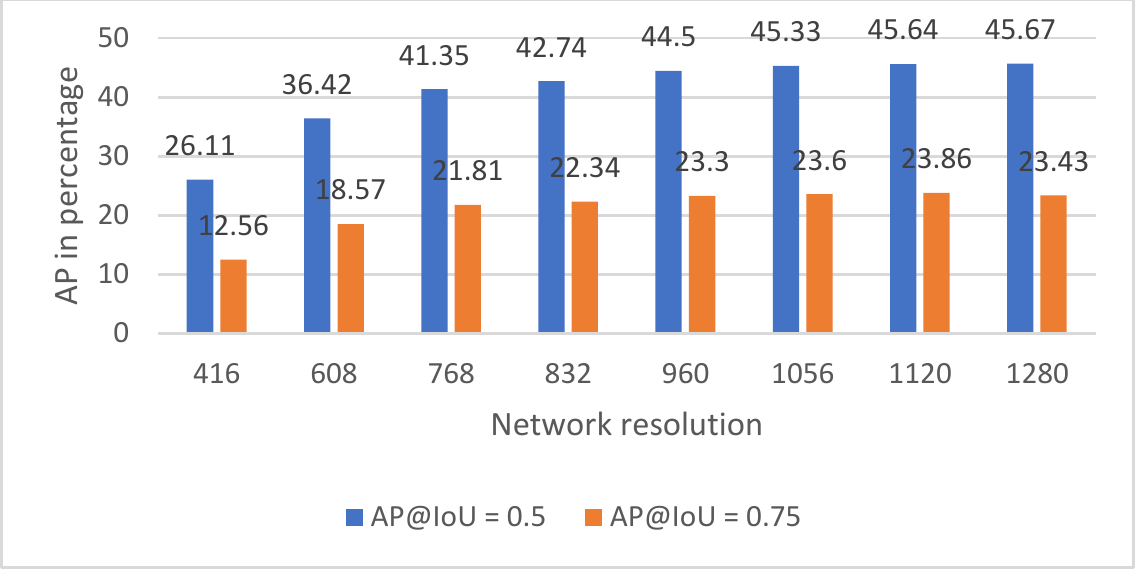}
   \caption{AP versus Network resolution}

  \end{minipage}
  \hfill
  \begin{minipage}[b]{0.49\textwidth}
    \centering
   \includegraphics[width=0.99\linewidth]{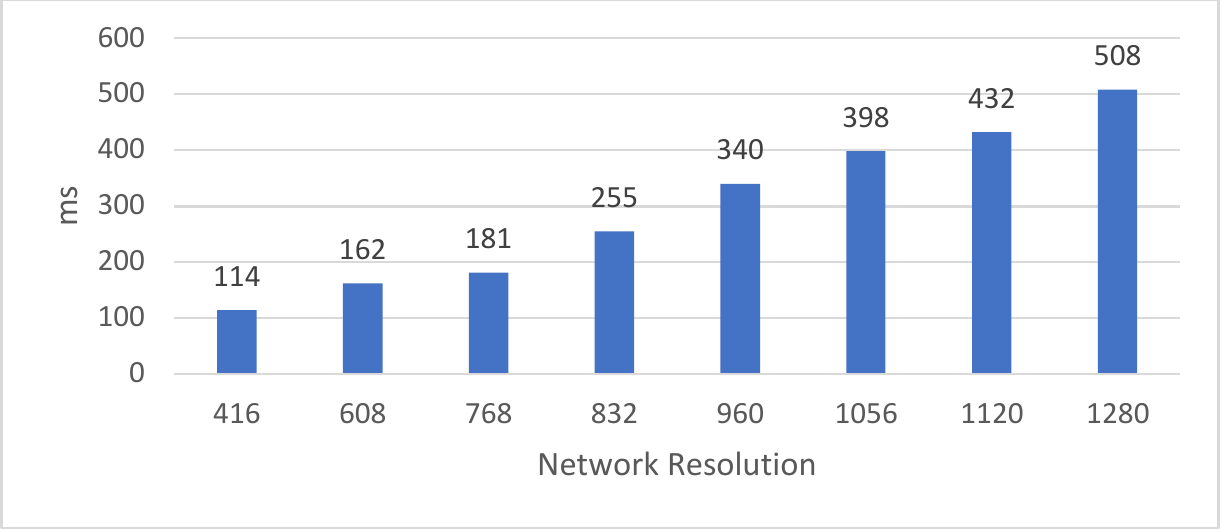}
   \caption{Inference time versus Network resolution on Tesla K80}
%
  \end{minipage}
\end{figure}
%
%
%
Our results inferred that the model performed best in terms of AP with a network size of 1120 * 1120. The model with 416 * 416 resolution utilized a minimum inference time of 114 ms as observed in Figure 8. It was then concluded that for all metric calculations, weights of the trained model at 8000 iterations would be loaded on the model with a network size of 1120 * 1120.
\subsubsection{Model Evaluation Results:}
The trained model is evaluated for VisDrone testset-dev using standard performance metrics. We have calculated precision for each object class for 0.5 IoU and the results are shown in Figure 9. The precision versus recall graph for varying IoU and confidence thresholds is shown in Figure 10. 
 \begin{figure}[h!]
 \centering
 \includegraphics[width=0.85\linewidth]{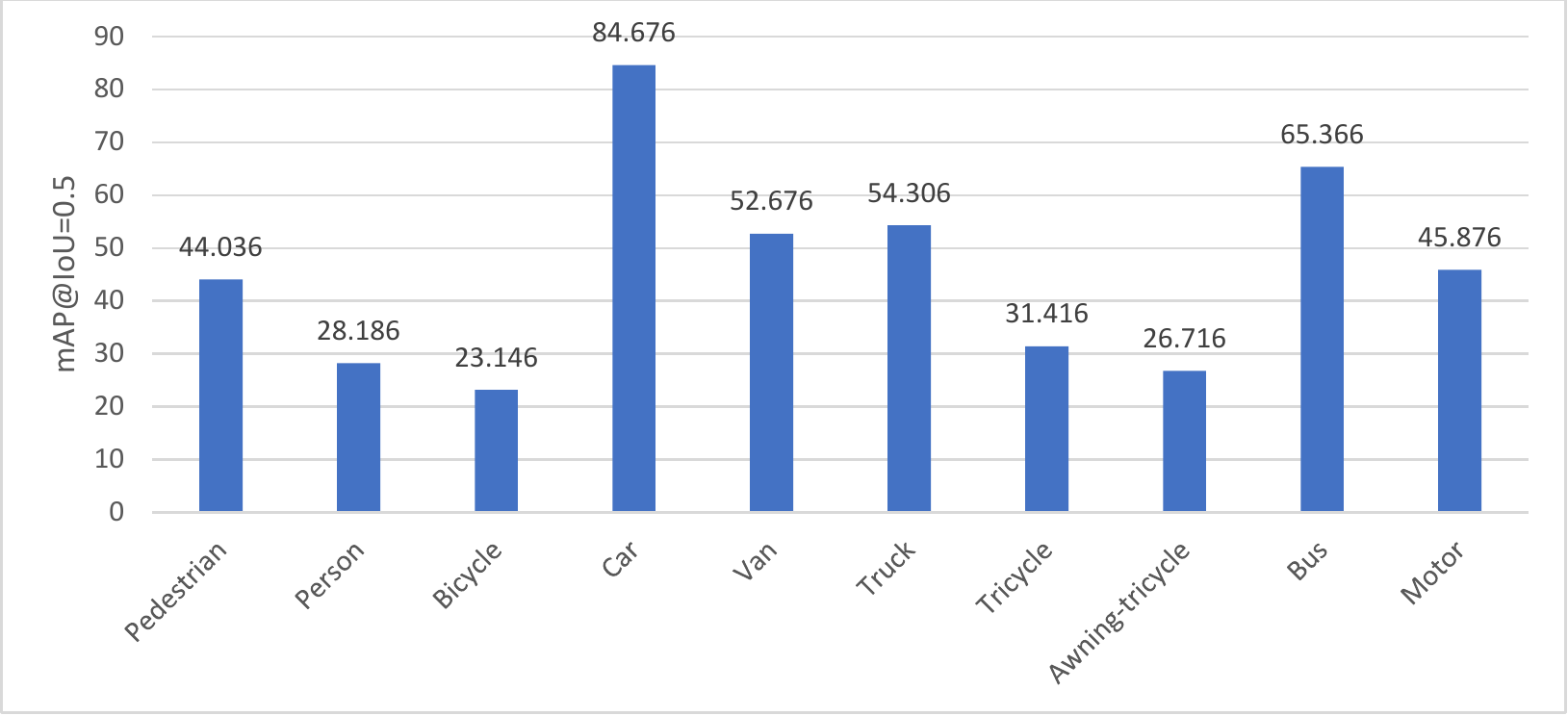}
 \caption{Object classwise performance}
%
 
 \end{figure}
\begin{figure}[h!]
  \centering
 \includegraphics[width=0.85\linewidth]{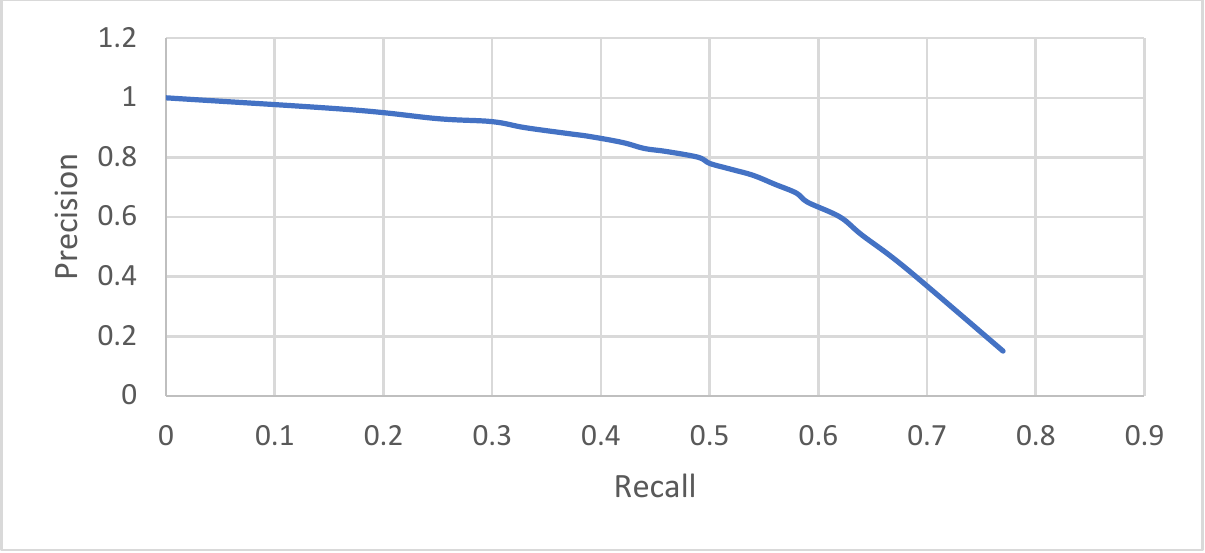}
 \caption{Precision versus recall curve}
  
  \end{figure}
 The model performed best on transport vehicles, which are in general larger in an image frame. It has the highest precision for `Car' (AP@0.5 = 84.676) followed by `Bus' (AP@0.5 = 65.366).
\subsection{Comparison with other Aerial Object Detectors}
We have compared our results with those obtained by state-of-the-art aerial object detectors tested on the VisDrone dataset and are explained below.
\subsubsection{Comparison with YOLOv3:}
 For a more relevant, useful, and comprehensive comparison, we have compared our results with the metrics obtained by Slim YOLOv3 \cite{slim}. As YOLOv3-SPP models have resulted in a higher mAP, we have compared our results only with them. The results are shown in Figure 11. It needs to be taken into account that the authors of the Slim YOLOv3 paper provided the results on the VisDrone validation set, and our results are for Visdrone testset-dev. However, it can be generalized that the results would have been similar in the test set-dev, as the images in both datasets are pretty identical.
  \begin{figure}[h!]
  \centering
  \includegraphics[width=0.85\linewidth]{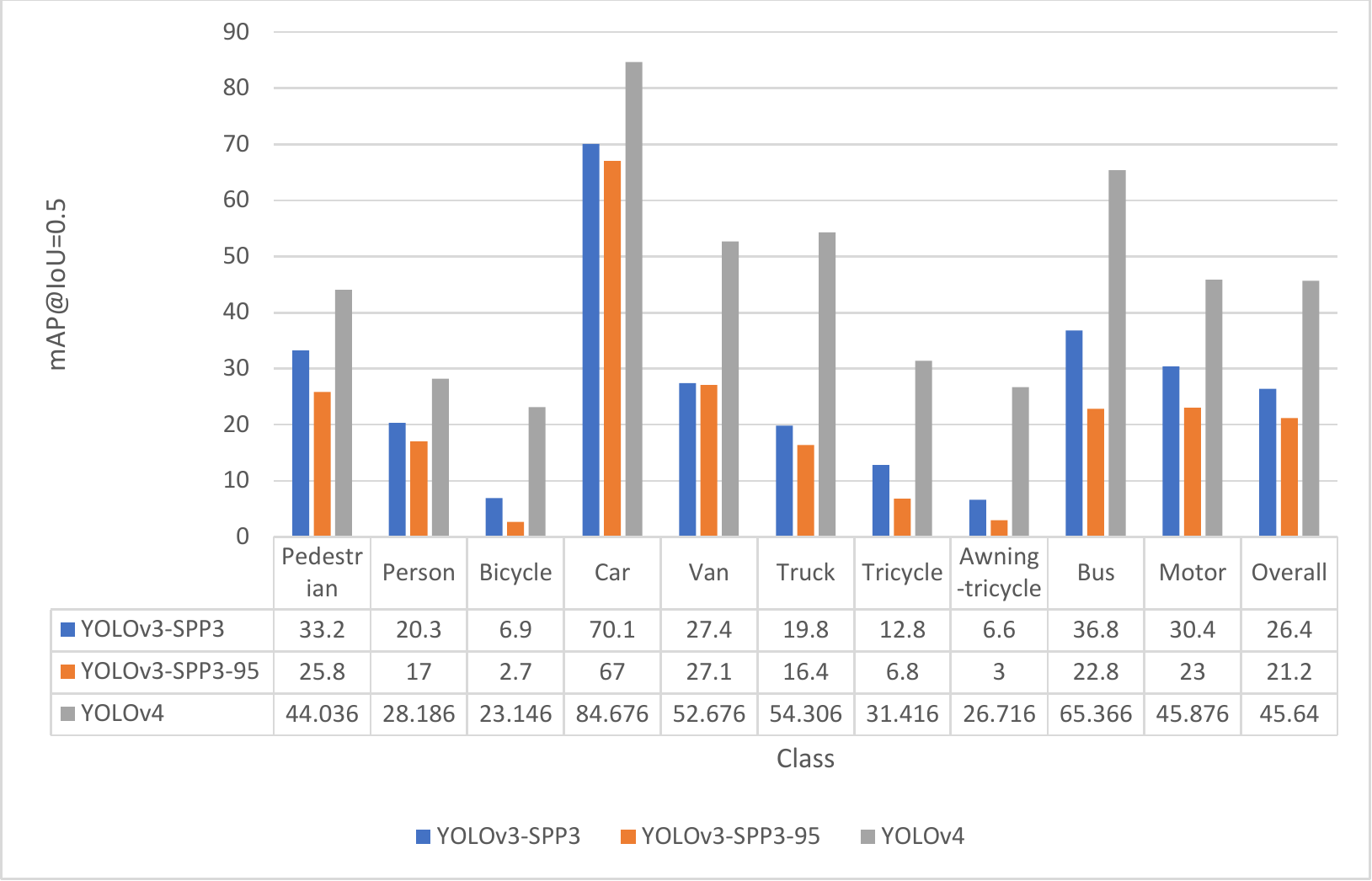}
  \caption{YOLOv4 versus YOLOv3 performance for aerial object detection}

  \end{figure}
  We can observe a huge improvement in the mAP performance for each object category by using YOLOv4, and it outperformed YOLOv3 by a notable amount.
 \subsubsection{Comparison with SOTA detectors from the VisDrone challenge:} The performance of the YOLOv4 algorithm is compared with baseline and best performing models of the VisDrone object detection challenge. The VisDrone2019-DET-toolkit is used for evaluation \cite{zhu2020vision}. The toolkit uses AP, AP@IOU=0.50, AP@IOU=0.75, ARmax=1, ARmax=10, ARmax=100, and ARmax=500 metrics to evaluate the results of detection algorithms. Unless otherwise specified, the AP and AR metrics are averaged over multiple IoU values. Specifically, the toolkit uses ten IoU thresholds of 0.50:0.05:0.95. Here, we would like to highlight that the VisDrone evaluation scheme did not consider occluded and truncated objects, whereas our method has considered these two categories. Table \ref{tab:1} shows the comparison of performance with VisDrone baseline detectors.
 
%
%
%
\begin{table}[!h]\small
     \caption{A comparison with models used in the VisDrone 2019 challenge \cite{pailla2019visdrone}. Models marked with '*' placed first, second and third respectively. Text with bold font highlights the best value for a metric.}
    \label{tab:1}
    \setlength{\tabcolsep}{6pt}
    \begin{tabular}{c|c|c|c|c|c|c|c}
     Methods & mAP &	AP50 &	AP75 &	AR1	& AR10 & AR100 & AR500  \\ \hline
     RetinaNet &	11.81 &	21.37 &	11.62 &	0.21 &	1.21 &	5.31 &	19.29  \\
      \hline
DetNet59 &	15.26 &	29.23 &	14.34 & 0.26 &	2.57 &	20.87 &	22.28   \\
 \hline
 Cascade-RCNN &	16.09 &	16.09 &	15.01 &	0.28 &	2.79 &	21.37 &	28.43 \\
 \hline
 CornerNet &	17.41 &	34.12 &	15.78 &	0.39 &	3.32 &	24.37 &	26.11 \\
 \hline
 FPN &	16.51 &	32.2 &	14.91 &	0.33 &	3.03 &	20.72 &	24.93  \\
 \hline
 Light-RCNN &	16.53 &	32.78 &	15.13 &	0.35 &	3.16 &	23.09 &	25.07 \\
 \hline
 DPNet-Ensemble* & \textbf{29.62} & 54 & \textbf{28.7} & 0.58 & 3.69 &17.10 & 42.37   \\
 \hline
 ACM-OD* &	29.13 &	54.07 &	27.38 &	0.32	 & 1.48	 & 9.46 &	44.53  \\
 \hline
 RRNet* & 29.13 &\textbf{55.82} &	27.23 &	1.02 &	8.5	& \textbf{35.19} &	\textbf{46.05} \\
 \hline
 							
 YOLOv4 &	18.5 &	35.72 &	17.93 &	\textbf{1.8} &	\textbf{13.4} &	30.11 &	30.59  \\
 \hline
    \end{tabular}
\end{table}

%
%
YOLOv4 performed better than most of the participating single as well as double stage detectors. Furthermore, our method achieved the highest AR, given 1 and 10 detections per image, than all other methods. The drastic difference between AP of the various object classes can be attributed to the unequal distribution of particular class instances in the dataset. It can further be added that the model worked well on all classes and achieved an AP value for each class corresponding to its number of instances present in the dataset. The comparison graph showing the inference speed of various algorithms is shown in Figure 12. 
\begin{figure}[!h]
  \centering
\includegraphics[width=0.85\linewidth]{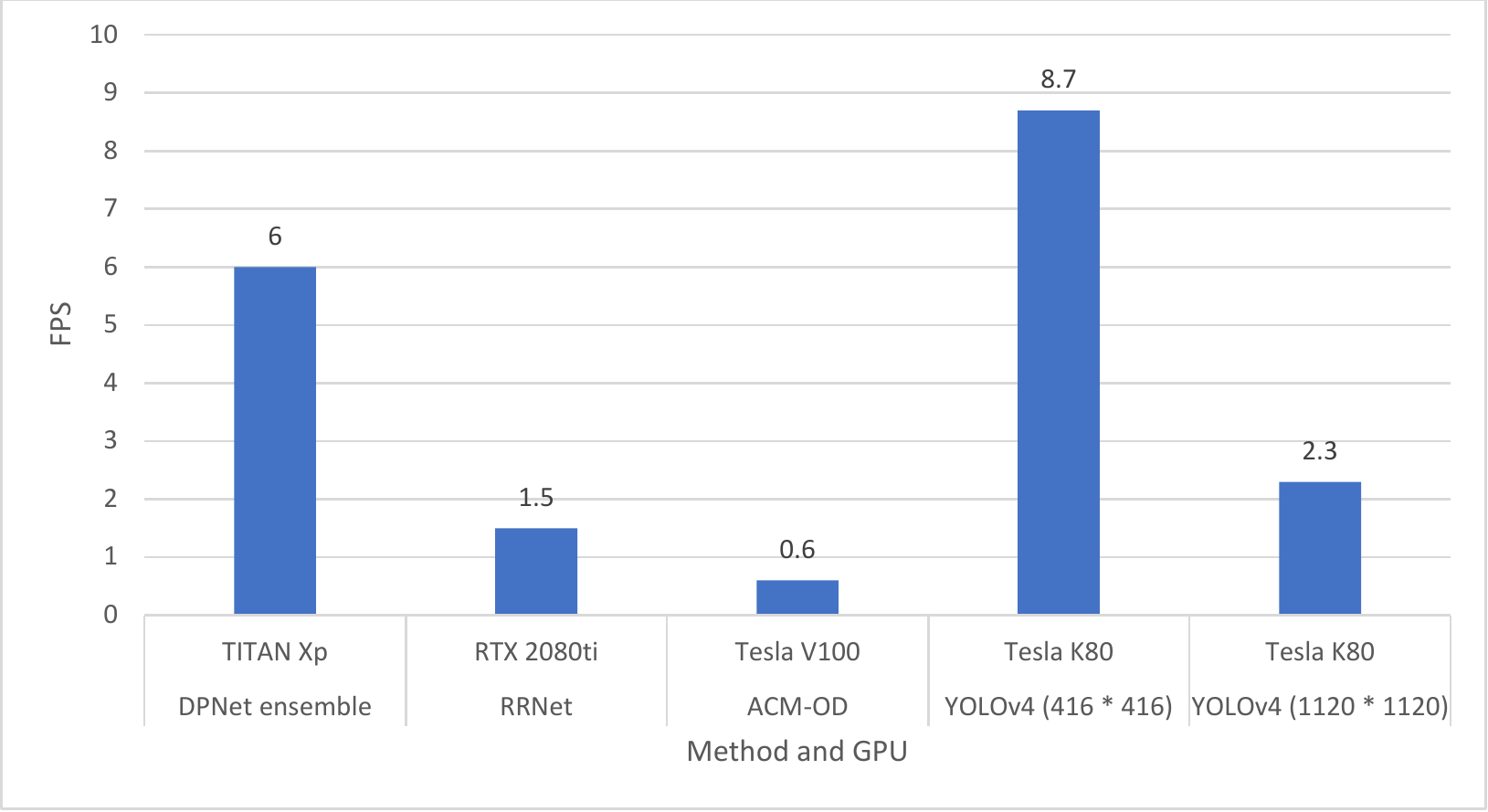}
\caption{Inference speed versus methods\cite{pailla2019visdrone}}
%
%
   \end{figure}

The maximum detection speed achieved with YOLOv4 on the VisDrone dataset was 8.7 FPS which is better when compared with the state-of-the-art detectors. ACM-OD and DPNet-ensemble are both two-stage detectors based on RCNN. As such detectors have discretized functions of classification and localization, they are significantly slower than single-stage detectors like YOLOv4, which are optimal for real-time detections tasks where hardware performance is limited. For similar reasons, RRNet, a hybrid model of the anchor-free detector and the two-stage detector, is slower than YOLOv4. The sample detections performed on images from the test set-dev dataset are shown in Figure 13.
   \begin{figure}[!t]
    \centering
     \includegraphics[width=0.85\linewidth]{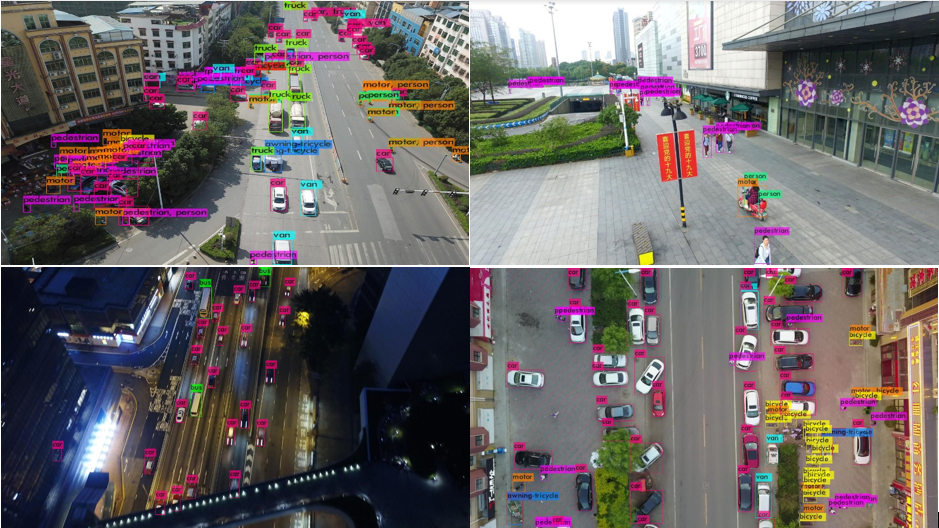}
     \caption{Object detection using YOLOv4 on the test-dev dataset}
%
     \end{figure}
    We have shown our model's capability in detecting the truncated and occluded objects in Figure 14.
   \begin{figure}[h!]
     \centering
 \includegraphics[width=0.85\linewidth]{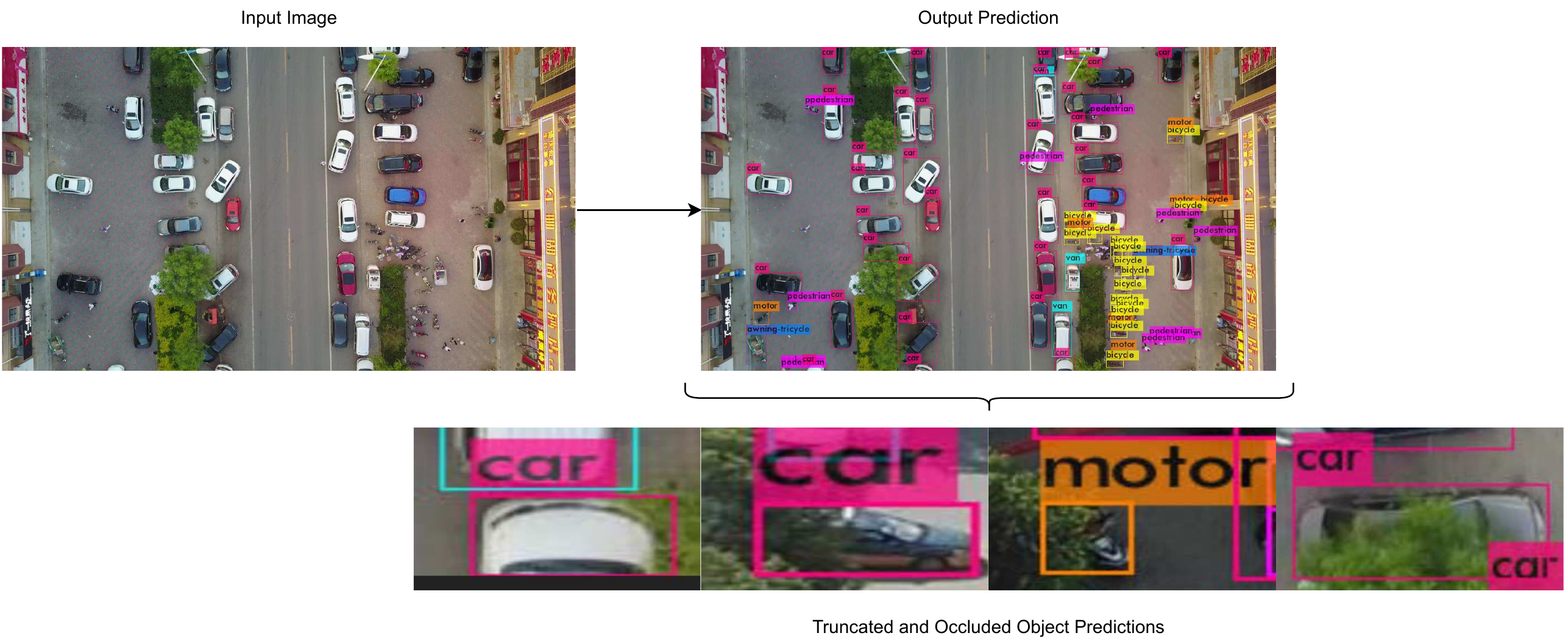}
 \caption{Original image and predicted output highlighting detection of truncated and occluded objects}
 
%
%
%
      \end{figure}

\subsection{Discussion} \label{sec:result2}

We initially tried two additional YOLO configurations besides the one described in the results. 
   \begin{enumerate}
  \item YOLOv4 architecture trained with a network resolution 416 * 416 resulted in an AP of 33.06\% (for 0.5 IoU) after 5000 iterations. However, the model did not perform well on the validation set as it was inaccurate in detecting smaller objects. The AP @0.50 for the `person' and `bicycle' classes was only 9.92\% and 8.91\%, respectively, after 5000 iterations. These metrics motivated us to increase the input network size from 416 * 416 to 768 * 768. 
  \item A model with the YOLOv3-5L architecture is a modification of YOLOv3 and consists of five YOLO layers. The network resolution was set as 416 * 416. The AP@0.50 for this model stagnated at 15.85\% after 5000 iterations. This was considerably less than YOLOv4 for the same number of iterations; therefore, this model was dropped.
  \end{enumerate}
  During our experimentation, we got the maximum inference speed of up to 8.7 fps for input resolution 416 * 416 and is reduced to 2.3 fps when we choose the input resolution of 1120 * 1120. Depending on the application, we can choose the optimum value of input resolution to have the right balance between AP and inference speed. YOLOv4 outperformed the YOLOv3 object detector in terms of detection performance. We compared our results with the best-performing approaches in the VisDrone object detection challenge. As our model has been trained to detect truncated objects as well (as shown in the Figure 14), the number of true positives decreases as the toolkit utilized for the evaluation of the challenge excludes predictions for objects with more than 50\% truncation ratio, which leads to lower precision and recall values for our model. Even then, YOLOv4 performed better than various detectors participating in the challenge in several metrics, and therefore it can be a better-suited algorithm for aerial platforms. Furthermore, the ability to detect truncated and occluded objects are highly favorable for aerial detection tasks as such instances regularly occur for images or videos shot from a drone.
  
  \par Using pre-trained weights enabled the model for better edge and pattern recognition, which is critical for object detection. This was done to facilitate transfer learning and accelerate the model's training and reduce convergence time.  
  
  \par Our experimental results reinforce the claims made by Alexey B et al. in his paper \cite{yolov4}. Through incorporating multiple innovative methods, a drastic increase in performance is observed when YOLOv4 is compared to YOLOv3. The implementation of SPP and PAN instead of the Field Pyramid Network (FPN) \cite{lin2017feature} used in YOLOv3 significantly increased the receptive field, separating the most critical context features with almost no reduction of the network operation speed. Apart from this, the inclusion of multiple data augmentation methods (Bag of Freebies) makes the model more robust and improves its accuracy for detecting the truncated and occluded objects.   
  

   
   \section{Conclusion}
   Object detection is a highly complex task for aerial datasets due to the abundance and density of small objects. The YOLOv4 architecture obtained optimal results in terms of both speed and accuracy on the VisDrone Aerial dataset. A single-stage object detection model with fast inference speed can be used in real-time object detection tasks. Hence, allowing even small drones to be equipped with high-level object detection technology. With the availability of computational resources, the model can be modified to have a  higher network resolution and size, further improving precision. Apart from this, training the model with higher resolution images can increase the accuracy even further.
%

\end{document}